\documentclass[runningheads]{llncs}

\usepackage{accv}

\usepackage{accvabbrv}

\usepackage{graphicx}
\usepackage{booktabs}
\usepackage{amsmath,amssymb} 
\usepackage{times}
\usepackage{epsfig}
\usepackage{graphicx}
\usepackage{amsmath}
\usepackage{amssymb}
\usepackage{booktabs}
\usepackage{multirow}
\usepackage{bbm}
\usepackage[ruled,linesnumbered]{algorithm2e}

\usepackage{fontawesome}

\usepackage{caption}

\usepackage[accsupp]{axessibility}  

\usepackage[pagebackref,breaklinks,colorlinks,citecolor=accvblue]{hyperref}

\usepackage{orcidlink}

\begin{document}

\title{Pluggable Style Representation Learning for Multi-Style Transfer} 

\titlerunning{Pluggable Style Representation Learning for Multi-Style Transfer}

\author{Hongda Liu\inst{1} \and
Longguang Wang\inst{2} \and
Weijun Guan\inst{1} \and
Ye Zhang\inst{1} \and
Yulan Guo\inst{1}
}

\authorrunning{Liu et al.}

\institute{The Shenzhen Campus of Sun Yat-Sen University, Sun Yat-Sen University \and 
Aviation University of Air Force \\
\email{\{liuhd36@mail2.sysu,guoyulan@sysu\}.edu.cn}}

\maketitle

\begin{abstract}
Due to the high diversity of image styles, the scalability to various styles plays a critical role in real-world applications. To accommodate a large amount of styles, previous multi-style transfer approaches rely on enlarging the model size while arbitrary-style transfer methods utilize heavy backbones. However, the additional computational cost introduced by more model parameters hinders these methods to be deployed on resource-limited devices. To address this challenge, in this paper, we develop a style transfer framework by decoupling the style modeling and transferring. Specifically, for style modeling, we propose a style representation learning scheme to encode the style information into a compact representation. Then, for style transferring, we develop a style-aware multi-style transfer network (SaMST) to adapt to diverse styles using pluggable style representations. In this way, our framework is able to accommodate diverse image styles in the learned style representations without introducing additional overhead during inference, thereby maintaining efficiency. Experiments show that our style representation can extract accurate style information. Moreover, qualitative and quantitative results demonstrate that our method achieves state-of-the-art performance in terms of both accuracy and efficiency. The codes are available in \href{https://github.com/The-Learning-And-Vision-Atelier-LAVA/SaMST}{https://github.com/The-Learning-And-Vision-Atelier-LAVA/SaMST}.
\end{abstract}

\section{Introduction}

\begin{figure}[t]
		\centering
    \includegraphics[width=0.8\linewidth]{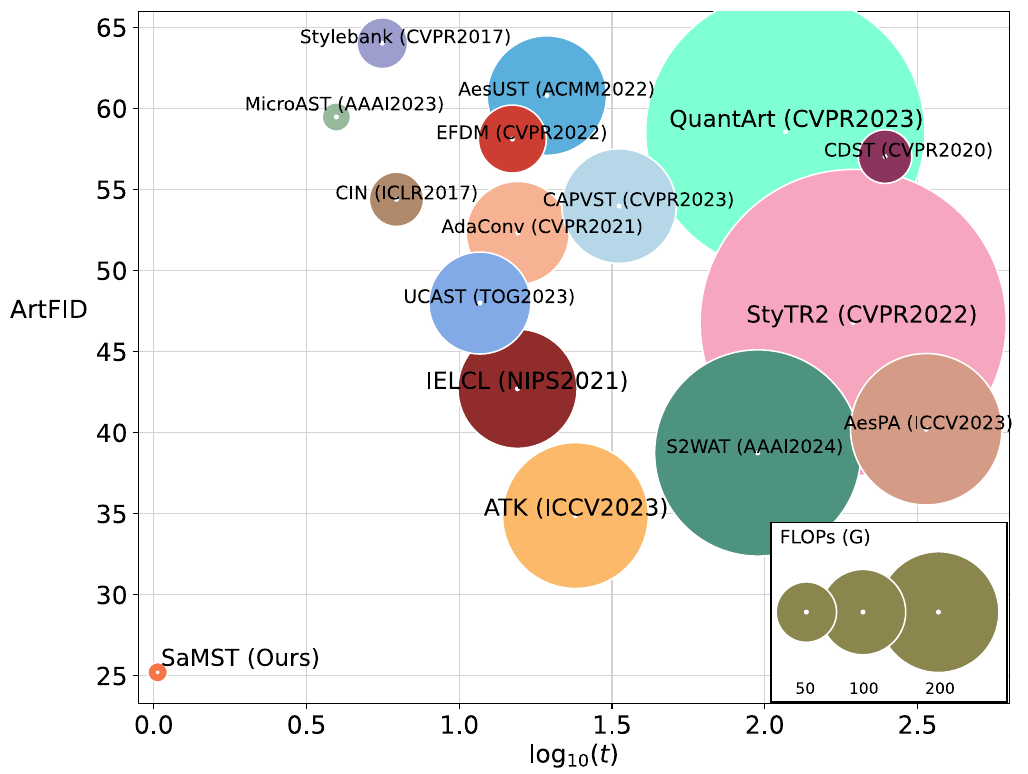}
		\caption{Trade-off between inference time $t$ (ms) and ArtFID~\cite{wright2022artfid} achieved by different methods. The size of a circle represents FLOPs.}
    \vspace{-15pt}
 \label{scatter1}
\end{figure}

Style transfer (ST) aims at capturing image style to generate artistic images, which has attracted increasing interests since the seminal works~\cite{gatys2015neural,gatys2016image}. Style extracting ability and generation quality are two critical research directions in the area of style transfer. In early years, a single model only accomplishes single style transfer (SST)~\cite{johnson2016perceptual,ulyanov2016texture,ulyanov2016instance}. To improve the flexibility of ST, multi-style transfer (MST) methods aim to incorporate multiple styles into one single model~\cite{chen2017stylebank,zhang2018multi,dumoulin2016learned}. Recently, arbitrary-style transfer (AST) methods are proposed for wider style domain~\cite{huang2017arbitrary,sheng2018avatar,li2019learning,deng2020arbitrary,chandran2021adaptive,zhang2022exact,zhu2023all,hong2023aespa,zhang2024s2wat,jing2020dynamic}.

Under sufficient computational resources, AST methods are able to handle a wide style domain. However, in real applications, deploying ST models on edge devices is highly demanded. Existing AST methods commonly suffer large model sizes and low inference efficiency, which hinders them to be deployed on resource-limited devices. To remedy this, several efforts are made to develop lightweight network structures \cite{shen2018neural,jing2020dynamic,wang2023microast}. However, the limited model capacity of lightweight models hinders them to accommodate diverse styles, resulting inferior quality of generated images. To strike a balance between inference efficiency and generation quality, several methods~\cite{chen2017stylebank,zhang2018multi,dumoulin2016learned,yanai2017conditional} are developed to store image styles in separate network structures. Although these methods achieve faster inference speed, their model size increases dramatically as style numbers increase, resulting in a huge storage burden. Moreover, these methods suffer from difficulties in extending to new styles~\cite{zhang2018multi,dumoulin2016learned,chen2017stylebank}.

\begin{figure}[t]
		\centering
\includegraphics[width=1\linewidth]{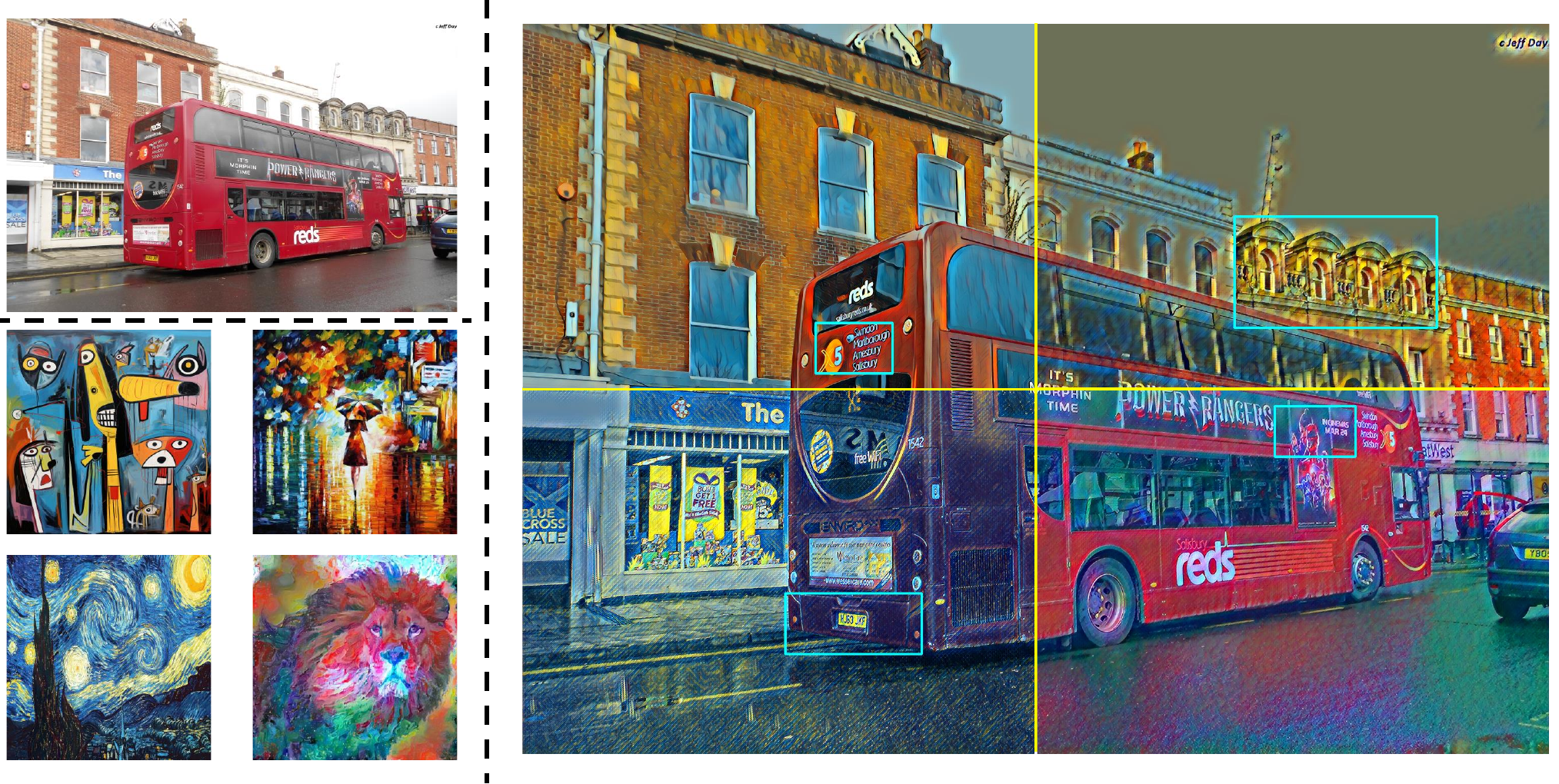}
  \vspace{-10pt}
		\caption{An 2K stylized sample ($2028\times1440$), rendered in about $0.01$ seconds on a single NVIDIA RTX 3090 GPU. The upper left and down left images are the content and style images, respectively. }
    \vspace{-10pt}
 \label{teaserfig}
\end{figure}

To address the issues above, we introduce a pluggable style representation learning scheme. This scheme inherits the high inference efficiency and generation quality of MST while maintaining superior style capacity of AST. Specifically, we encode the style-specific information into a compact representation and store it in a style codebook (SCB). Moreover, we propose a style-aware multi-style transfer (SaMST) network with flexible adaption to different styles based on the learned representations. Particularly, our SaMST incorporates style information to perform feature adaption by predicting convolutional kernels, variance/mean values and channel-wise modulation coefficients from the compact style representation. Our SaMST can reduce the model size and computational complexity while improving the quality of results, striking a better balance between accuracy and efficiency, as illustrated in Fig.~\ref{scatter1}. In addition, we propose an incremental style extension scheme. This scheme enables our model to quickly adapt to a new style representation without forgetting the previous styles. Experiments show our SaMST not only produces visually more pleasing results (Fig.~\ref{teaserfig} and Fig.~\ref{fig_2K}), but also achieves over $4\times$ reduction in model size and over $3\times$ speedup for each style (Table~\ref{comparisonwsotas}). 

In summary, our contributions are three-fold:
\begin{itemize}
    \item We introduce a pluggable style representation learning scheme for MST. Moreover, we propose a style-aware multi-style transfer (SaMST) network with flexible adaption to different styles based on the learned representations. And our SaMST achieves great advantages in terms of efficiency.
    \item To solve the size explosion and style catastrophic forgetting of previous MST methods, we propose a novel style representation extension scheme which has stronger application advantages.
    \item Extensive experiments show that our method produces state-of-the-art results in terms of both visual quality and quantitative performance.
\end{itemize}

\section{Related Work}

\subsection{Neural Style Transfer}

In earlier stage, Gatys \emph{et al.}~\cite{gatys2015neural,gatys2016image} propose optimization-based methods to obtain stylized images. However, numerous iterations are required to obtain a satisfied result. To achieve faster generation speed, feed-forward methods~\cite{johnson2016perceptual,ulyanov2016texture,ulyanov2016instance} are proposed. They force the whole network to learn a certain style, so that it achieves fast style transfer with any content image. Furthermore, researchers adapt multiple image styles to corresponding network structures \cite{chen2017stylebank,zhang2018multi,dumoulin2016learned} to enhance the generalization of the ST. Some researchers find that pre-trained VGG~\cite{simonyan2014very} can accurately capture image content and style information. So it was applied to style transfer as an image feature encoder. The scheme is capable of any image style~\cite{huang2017arbitrary,sheng2018avatar,li2019learning,deng2020arbitrary,chandran2021adaptive,zhang2022exact,zhu2023all,hong2023aespa,jing2020dynamic}. With the wide application of attention mechanism and transformer structure in computer vision, some researchers utilize them to enhance stylized image quality~\cite{zhang2024s2wat,deng2022stytr2,wu2021styleformer}. Inspired by the developments of contrastive learning, several efforts~\cite{chen2021artistic,zhang2023unified,wang2023microast,zhang2022domain} have been made to leverage contrastive learning to obtain better stylized results. In addition, as a practical image generation technique, the quality of style transfer~\cite{wright2022artfid,wang2021evaluate,yeh2020improving}, the diversity of style textures~\cite{li2017diversified,ulyanov2017improved,wang2020diversified}, user control~\cite{champandard2016semantic,gatys2017controlling,kolkin2019style} and other practical issues also attract researchers' concern.

In real applications, model efficiency and size are crucial. As the seminal work of feed-forward methods, \cite{johnson2016perceptual} greatly improves the inference speed compared to optimization based methods. Doumoulin \emph{et al.}~\cite{dumoulin2016learned} stores style-specific parameters into a learnable instance norm layer to reduce model size. Chen \emph{et al.}~\cite{chen2017stylebank} proposes stylebank to store a style in a certain module. Specific style module reduces inference time, at the cost of storage. MFS methods suffer from the problem of extreme model size inflation~\cite{chen2017stylebank} or style catastrophic forgetting~\cite{dumoulin2016learned,zhang2018multi} when finetuning new styles. Researchers further design lightweight models~\cite{shen2018neural,jing2020dynamic,wang2023microast} based on AST. With the developments of knowledge distillation\cite{hinton2015distilling}, some methods compress the streamlined ST models~\cite{wang2020collaborative,chiu2022pca} from large pre-trained models. However, AST methods still face the problems of model redundancy, slow inference and weak perception in wider style domain.

\subsection{Efficient Network Architecture}

Inspired by prior research on transformation-invariant scattering~\cite{sifre2013rotation}, Laurent Sifre develops depthwise convolutions, and uses them in AlexNet to obtain small gains in accuracy and large gains in convergence speed, as well as a significant reduction in model size~\cite{vanhoucke2014learning}. Then this designed layer is used in many classic efficient vision backbones (\emph{e.g.},  Inception V1 and V2~\cite{szegedy2015going,ioffe2015batch}, Xception~\cite{chollet2017xception} and MobileNets~\cite{howard2017mobilenets}). In recent years, depthwise convolutions are applied in some practical tasks to improve inference efficiency, such as image super-resolution~\cite{wang2021unsupervised,xia2022knowledge}, neural machine translation~\cite{kaiser2017depthwise} and style transfer~\cite{chandran2021adaptive}. Besides, Jin~\emph{et.al}~\cite{jin2014flattened} also propose a flattened network that consist of consecutive sequence of one-dimensional filters across all directions in 3D space, which obtains comparable performance as conventional convolutional networks and reduces model size. Wang~\emph{et.al}~\cite{wang2017factorized} propose to factorize the convolutional layer to reduce its computation.

\section{Methodology}

\subsection{Overview}
\label{overview_sec}

\begin{figure*}[t]
		\centering
    \includegraphics[width=1\linewidth]{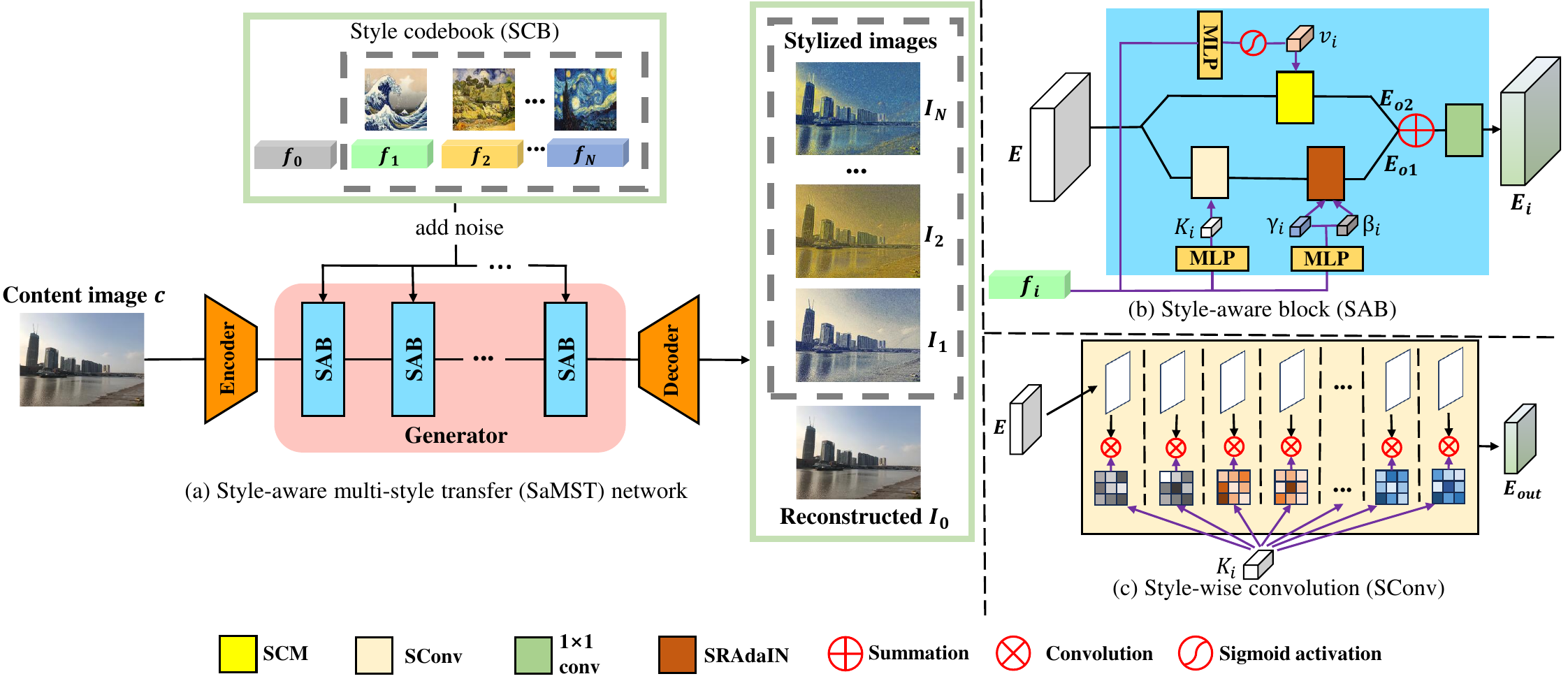}
		\caption{An overview of our multi-style transfer framework.}
    \vspace{-15pt}
 \label{mainnet}
\end{figure*}

Our multi-style style transfer framework consists of style-aware multi-transfer (SaMST) network and style codebook (SCB), as shown in Fig.~\ref{mainnet}(a). First, the content image $c$ is fed to Encoder to obtain content feature $E$. Then, a style representation (\emph{e.g.}, $f_i\in{\mathbb{R}}^{{C}}$ and $C=16$) is selected as style condition information from SCB, which is employed to adapt the generator parameters. Next, $E$ is fed to the generator to obtain stylized image feature $E_{i}$. Finally, $E_{i}$ is decoded by the decoder to obtain stylized image $I_i$. Note that, our framework is compatible to diverse  encoder and decoder architectures. For real applications (\emph{i.e.}, efficiency and model size), a lightweight 3-layer symmetric encoder and decoder is employed in our framework, which is similar to \cite{chen2017stylebank,dumoulin2016learned,xu2019styleremix}.

\subsection{Style-aware Multi-style Transfer (SaMST) Network}

\subsubsection{Style-wise Convolution (SConv)}

To better preserve local geometric structures of style images, AdaConv~\cite{chandran2021adaptive} employs a style-aware depthwise convolution. Inspired by AdaConv, our SConv (Fig.~\ref{mainnet}(c)) layer learns to predict the kernel of a depthwise convolution conditioned on a style representation to achieve flexible style-aware adaption. Our SConv not only makes the network lightweight and flexible, but also captures more local geometric structures to promote stylized results.

Specifically, we obtain a style representation by explicitly selecting from the provided image styles (\emph{e.g.}, $f_i$). Then, this representation is used as conditional information and passed to a MLP to generate the convolution kernels $K_i$ in the style-aware block (Fig.~\ref{mainnet}(b)). Note that, $K_i \in {\mathbb{R}}^{C_{in}\times{1}\times{k_w}\times{k_h}}$ and $C_{in}$ represents channel number of the input content image feature $E$. Details of the style-wise convolution are shown in Fig.~\ref{mainnet}(c). The predicted convolution kernels perform depthwise convolution operation on content image feature $E$. Afterwards, the style-wise convolution is conducted as follows:
\begin{align}
\label{groupconv}
E_{out} = {\rm Sconv}(K_i, E; groups)=K_i{\otimes}E,
\end{align}
where $groups$ is the number of convolution groups, and $\otimes$ denotes the depthwise convolution operation.

\subsubsection{Style-representation Adaptive Instance Norm (SRAdaIN)}

In addition to local geometric structures, the global properties are also critical to the final results. Inspired by \cite{huang2017arbitrary}, we propose SRAdaIN to further capture global properties from style images. Specifically, the style representation $f_i$ is fed to a MLP to predict the mean $\gamma_i\in {\mathbb{R}}^{C_{in}}$ and variance $\beta_i\in {\mathbb{R}}^{C_{in}}$ of the style. Afterwards, SRAdaIN is performed as follows:
\begin{align}
\label{radain}
E_{o1} = {\rm SRAdaIN}(E_{out},\gamma_i,\beta_i)={\beta_i}\Big{(}\frac{E_{out}-\mu(E_{out})}{\sigma(E_{out})}\Big{)} + \gamma_i,
\end{align}
where $\mu(\cdot), \sigma(\cdot)$ refer to compute mean and variance of input features, respectively, which is similar to~\cite{dumoulin2016learned,huang2017arbitrary}.

\subsubsection{Style-wise Channel Modulation (SCM):} Inspired by CResMD~\cite{he2020interactive} that uses controlling variables to rescale different channels to handle multiple image degradations, our SCM learns to generate modulation coefficients based on the style representation to perform channel-wise feature adaption. Specifically, the selected style representation $f_i$ is passed to another MLP and a sigmoid activation layer to generate channel-wise modulation coefficients $v_i$. Then, $v_i$ is used to rescale different channel components in $E$, obtaining $E_{o2}$.
\begin{align}
E_{o2} = {\rm SCM}(E,v_i)=E\odot{v_i},
\end{align}
where $\odot$ refers to element-wise multiplication with broadcast over the spatial dimensions. Finally, $E_{o1}$ is summed up with $E_{o2}$ and fed to the subsequent layers to produces the stylized output feature $E_{i}$.

\subsubsection{Discussion:} Previous AST methods rely on heavy backbones to extract style information from style images at the cost of high computational overhead and long inference time. In contrast, we encode local geometric structures and global properties from style images into compact representations (16-dimension). Our method significantly reduces the computational complexity and produces notable inference speedup. Besides, in contrast to heavy style extractors, the pluggable style representation contains sufficient style information in a very small storage space. It quickly expands to model parameters, which greatly reduces the model size and storage burden.

\subsection{Framework Training}

\subsubsection{Training Loss}
The overall loss function consists of a content term, a style term, a reconstruction term, and a geometric term, which is defined as follows:
\begin{align}
\label{loss_fun}
    \mathcal{L}=\lambda_{c}\mathcal{L}_{c}(I_i,c) + \lambda_{s}\mathcal{L}_{s}(I_i,s_i)+\lambda_{ae}\mathcal{L}_{ae}(I_0,c)+\lambda_{geo}\mathcal{L}_{geo}(c,i),    
\end{align}
where $\lambda_{c}$, $\lambda_{s}$, $\lambda_{ae}$ and $\lambda_{geo}$ are set to $1$,$10$, $0.01$ and $0.01$, respectively. We show the training process in Algorithm~\ref{algorithm_training}.

\noindent\textbf{(1) Content loss and style loss:}

Similar to previous works \cite{deng2022stytr2,zhang2024s2wat,deng2021arbitrary}, we define content and style loss as follows:
\begin{align}
\mathcal{L}_{c}(I_i,c)=\sum_{l \in \{l_c\}}||VGG^l(I_i)-VGG^l(c)||_2,
\end{align}
\begin{align}
\mathcal{L}_{s}(I_i,s_i)=\sum_{l \in \{l_s\}}(||\mu(VGG^l(I_i))-\mu(VGG^l(s_i))||_2+ \notag \\
||\sigma(VGG^l(I_i))-\sigma(VGG^l(s_i))||_2),
\end{align}
where $VGG^l$ refers to features extracted from the $l$-th layer in a pre-trained VGG-16~\cite{simonyan2014very}.  $\mu(\cdot)$ and $\sigma(\cdot)$ denote the mean and variance of extracted features, respectively.

\noindent\textbf{(2) Reconstruction loss:}

In order to allow users to edit stylization degree, a style representation $f_0$ is set to represent the content and style information of the content image itself (\emph{i.e.}, a set of auto-encoder parameters is stored in $f_0$). The reconstruction loss is as follows:
\begin{align}
    \mathcal{L}_{ae}(I_0,c)=||I_0-c||_2.
\end{align}

\noindent\textbf{(3) Geometric loss:}

Geometric consistency, which is introduced in recent works~\cite{fu2019geometry,maeda2020unpaired}, reduces the space of possible translation to preserve the scene geometry. Inspired by geometric consistency, we implement geometric consistency loss to promote stylized results.
\begin{align}
    \mathcal{L}_{geo}(c,i)=\sum_{t \in \{T\}}(&|| {\rm SaMST}(c,f_i)-t^{-1}({\rm SaMST}(t(c),f_i)) ||_1 
\notag \\
+ & || {\rm SaMST}(t(c),f_i)-t({\rm SaMST}(c,f_i)) ||_1),
\end{align}
where $\{T\}$ represent 8 distinct patterns of flip and rotation.

\vspace{-20pt}
\begin{algorithm}
\caption{Pipeline to train a general model}\label{algorithm_training}

\SetKwInOut{Preparation}{Data}
\SetKwInOut{Target}{Target}
\SetKwInOut{Initial}{Initial}

\Preparation{content images, $N$ style images $\left\{s_1,s_2,...,s_N\right\}$ and corresponding style indices $\left\{1,2,...,N\right\}$.}
\Target{${\rm SCB}=\left\{f_0,f_1,f_2,...,f_N\right\}$, ${\rm SaMST}$.}
\Initial{$f\leftarrow{\mathbbm{1}}$ for every $f$ in ${\rm SCB}$, $iterations\leftarrow$ training iterations, $r \leftarrow 0$.}

\While{$r \leq iterations$}{

$\mathord{\bullet}$ Randomly sample $b$ content images $\mathit{C}=\{c_j\}$ and $b$ style indices $\mathit{Y}=\{y_j\}$ ($j \in \{1, ..., b\}$ and $y_j \in \{1,...,N\}$) as one mini-batch. According to $\mathit{Y}$, select corresponding style images $\mathit{S}=\{s_{y_j}\}$ and style representations $\mathit{F}=\{f_{y_j}\}$

$\mathord{\bullet}$ Inference: stylized images $\mathit{I}={\rm SaMST}(\mathit{C},\mathit{F})$ (\emph{i.e.}, $\mathit{I}=\{{c_j}{s}_{y_j}\}=\{{I_{y_j}}\}$); reconstructed content images $\mathit{I_0}={\rm SaMST}(\mathit{C},f_0)$

$\mathord{\bullet}$ Loss: $\mathcal{L}=\lambda_{c}\mathcal{L}_{c}(\mathit{I},\mathit{C})+\lambda_{s}\mathcal{L}_{s}(\mathit{I},\mathit{S})+\lambda_{ae}\mathcal{L}_{ae}(\mathit{I_0},\mathit{C})+\lambda_{geo}\mathcal{L}_{geo}(\mathit{C,Y})$

$\mathord{\bullet}$ update: ${\rm SaMST}$ and $(\mathit{F},f_0)$

$\mathord{\bullet}$ $r\leftarrow r+1$\;
}
\end{algorithm}
\vspace{-30pt}

\subsubsection{Incremental Training}
Previous MST methods need to finetune their models when extending new styles~\cite{zhang2018multi}, which results in style catastrophic forgetting. In addition, the model size of these methods increases as the number of styles increases. For example, Stylebank\cite{chen2017stylebank} stores styles in heavy network structures and requires $1.18M$ parameters for each new style (as shown in Table~\ref{comparisonwsotas}).

To address problems above, we propose a novel scheme for style extension. SaMST is trained on a large number of style images, and we believe that it already has strong generalization to address wider style domain. When adding new styles, we fix the SaMST and just update new style representations iteratively. We show the incremental training process in Algorithm~\ref{algorithm_incre_training}.

\begin{algorithm}
\caption{Incremental training}\label{algorithm_incre_training}

\SetKwInOut{Preparation}{Data}
\SetKwInOut{Fixed}{Fixed}
\SetKwInOut{Target}{Target}
\SetKwInOut{Initial}{Initial}

\Preparation{content images, M incremental style images $\left\{s_{N+1},s_{N+2},...,s_{N+M}\right\}$ and corresponding style indices $\left\{N+1,N+2,...,N+M\right\}$.}
\Fixed{Previous stable SCB: ${\rm SCB}_{fixed}=\left\{f_0,f_1,...f_N\right\}$, ${\rm SaMST}$.}
\Target{${\rm SCB}_{add}=\left\{f_{N+1},f_{N+2},...,f_{N+M}\right\}$.}
\Initial{$f\leftarrow{\mathbbm{1}}$ for every $f$ in ${\rm SCB}_{add}$, $iterations\leftarrow$ incremental training iterations, $r \leftarrow 0$.}

\While{$r \leq iterations$}{

$\mathord{\bullet}$ Randomly sample $b$ content images $\mathit{C}=\{c_j\}$ and $b$ style indices $\mathit{Y}=\{y_j\}$ ($j \in \{1, ..., b\}$ and $y_j \in \{N+1,...,N+M\}$) as one mini-batch. Then select corresponding style images $\mathit{S}=\{s_{y_j}\}$ and style representations $\mathit{F}=\{f_{y_j}\}$

$\mathord{\bullet}$ Inference: stylized images $\mathit{I}={\rm SaMST}(\mathit{C},\mathit{F})$ (\emph{i.e.}, $\mathit{I}=\{{c_j}{s}_{y_j}\}=\{{I_{y_j}}\}$).

$\mathord{\bullet}$ Loss: $\mathcal{L}=\lambda_{c}\mathcal{L}_{c}(\mathit{I},\mathit{C})+\lambda_{s}\mathcal{L}_{s}(\mathit{I},\mathit{S})+\lambda_{geo}\mathcal{L}_{geo}(\mathit{C,Y})$

$\mathord{\bullet}$ update: $\mathit{F}$

$\mathord{\bullet}$ $r\leftarrow r+1$\;

}
${\rm SCB}={\rm SCB}_{fixed}+{\rm SCB}_{add}$

\end{algorithm}

\subsubsection{Discussion:} Our SaMST differs from previous MST methods~~\cite{chen2017stylebank,dumoulin2016learned} in two aspects. 
(1) Architecture: Style-specific filters in Stylebank consist of vanilla convolution and instance norm layers ($1.18M$ parameters). In contrast, we store styles in much more compact representations ($16$ parameters). To utilize these representations, we design a universal and novel SaMST with SConv, SRAdaIN and SCM. (2) Loss: {As compared to widely used losses, we introduce an additional geometric loss to promote generation quality.}

\section{Experiments}

\subsection{Implementation Details}
We use MS-COCO~\cite{lin2014microsoft} as content dataset and select $50k$ style images from
WikiArt~\cite{phillips2011wiki} and the Internet. Style representation length $C$ in Sec.~\ref{overview_sec} is set to 16. And generator contains 3 SABs. During training, content images are rescaled to $256\times256$ pixels and style images are rescaled to $512\times512$ pixels. 8 content-style image patch pairs are randomly selected as a mini-batch. We adopt the Adam optimizer~\cite{kingma2014adam} to train the whole network for $3M$ iterations. The initial learning rate is set to 0.001 and decreased to half every $0.75M$ iterations. During test phase, SaMST can handle any input size as it is fully convolutional. More implementation details are available in supplemental material.

\vspace{-10pt}
\subsection{Comparison with Prior Arts}

We compare our SaMST to recent state-of-the-art ST methods, including CDST~\cite{wang2020collaborative}, AdaConv~\cite{chandran2021adaptive}, StyTR2~\cite{deng2022stytr2}, EFDM~\cite{zhang2022exact}, CAPVST~\cite{wen2023cap}, ATK~\cite{zhu2023all}, MicroAST~\cite{wang2023microast}, S2WAT~\cite{zhang2024s2wat} and Stylebank~\cite{chen2017stylebank}. Among them, Stylebank\footnote{Stylebank~\cite{chen2017stylebank} contains $1.18M$ parameters for each style, which results in heavy storage burden. So we just train stylebank on $500$ style images.} is a typical MST method while others are AST methods. We obtain the results of the methods by following their official code with default configurations. More comparison results with more methods are available in supplemental material.


\begin{figure*}[t]
		\centering
    \includegraphics[width=1\linewidth]{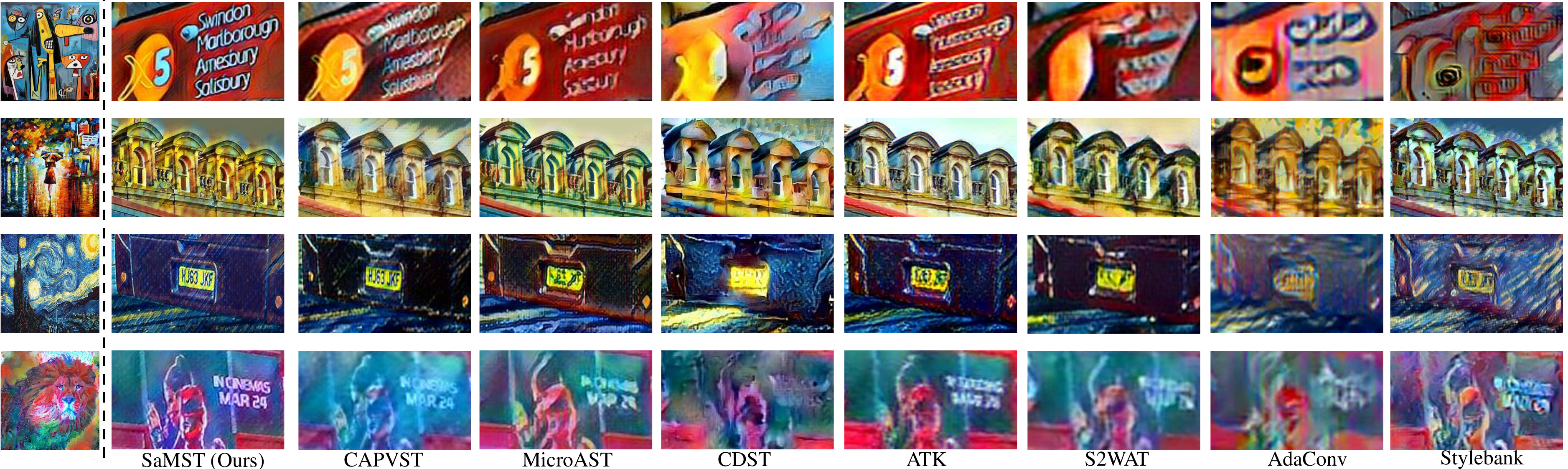}
    \vspace{-15pt}
    \caption{Visualization results of image details produced by different methods on a 2K image from Flickr2K dataset. The whole content image is shown in Fig.~\ref{teaserfig}.}
    \vspace{-5pt}
 \label{fig_2K}
\end{figure*}

\begin{figure*}[t]
		\centering
    \includegraphics[width=1\linewidth]{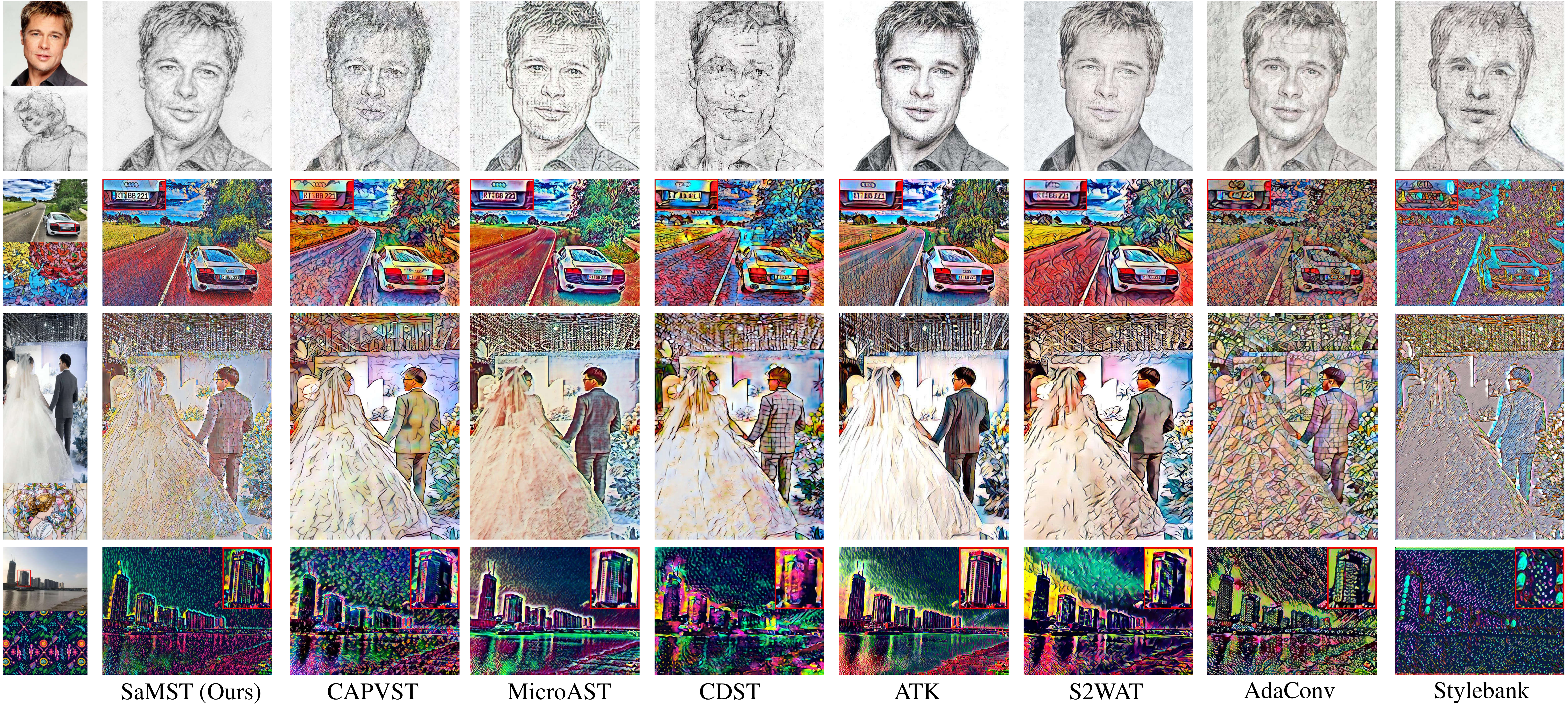}
    \vspace{-15pt}
    \caption{Qualitative comparison with the state of the art. Please zoom in for best view.}
    \vspace{-20pt}
 \label{fig_3}
\end{figure*}

\vspace{-15pt}
\subsubsection{Qualitative Comparison}

We show the visual comparisons in Fig.~\ref{fig_2K} and Fig.~\ref{fig_3}. In Fig.~\ref{fig_2K}, MicroAST~\cite{wang2023microast} and CDST~\cite{wang2020collaborative}, which are also designed as lightweight and fast models, suffer from severe performance drop. MicroAST produces compression artifacts (\emph{e.g.}, the $4^{th}$ row), while CDST breaks image local geometric structures and content details (\emph{e.g.}, the $1^{st}$ row). AdaConv~\cite{chandran2021adaptive} is proposed to capture more local structures from style images. As shown in the $1^{st}$ row, AdaConv focuses on local geometry of style images excessively, resulting in severe image distortion and artifacts. CAPVST~\cite{wen2023cap}, ATK~\cite{zhu2023all} and S2WAT~\cite{zhang2024s2wat} are also hard to achieve satisfied results. As for MST method, Stylebank~\cite{chen2017stylebank}  transfers insufficient style properties and destroys content details. In contrast, our SaMST produces more clear details and achieves higher perceptual quality (\emph{e.g.}, the texts in the $1^{st}$ row and license plate in the $3^{rd}$ row). And our SaMST captures sufficient colors and textures from style images, while it also keeps content structures accurately (\emph{e.g.}, the $2^{nd}$ row and the $4^{th}$ row).

As shown in Fig.~\ref{fig_3}, our SaMST achieves best visual quality to keep balance between style and content in the whole image. For example, our method captures detailed content information rather than other methods (\emph{e.g.}, $2^{nd}$ and $4^{th}$ row). Moreover, stylization generated by our SaMST are more closed to style images. In contrast, other methods suffer from content distortion (\emph{e.g.}, AdaConv) and insufficient stylization (\emph{e.g.}, Stylebank), and so on.

\begin{table*}[t]
		\caption{Quantitative comparison of the style transfer methods.Methods marked with ${^*}$ are MST approaches, while other methods are AST approaches. The \textbf{best} and \underline{second best} results are highlighted, respectively. Run time and FLOPs are evaluated on $512\times512$ images. "$+$" represents that the method expands new styles without forgetting. 'OIP' is short for 'once inference parameters', which refers to the number of parameters involved in one stylization inference for a certain style.}
  \vspace{-20pt}
        \renewcommand{\arraystretch}{1} 
		\begin{center}
			\small
                \resizebox{1\hsize}{!}{
				\begin{tabular}{|l|cccccccccc|}
					\hline 
					\multirow{1}{*}{Metric} 
                        
					&  \multirow{1}{*}{CDST~\cite{wang2020collaborative}} 
					& \multirow{1}{*}{AdaConv~\cite{chandran2021adaptive}} 
                        & \multirow{1}{*}{StyTR2~\cite{deng2022stytr2}} 
					&  \multirow{1}{*}{EFDM~\cite{zhang2022exact}} 
					& \multirow{1}{*}{CAPVST~\cite{wen2023cap}} 
                        & \multirow{1}{*}{ATK~\cite{zhu2023all}} 
					&  \multirow{1}{*}{MicroAST~\cite{wang2023microast}} 
					& \multirow{1}{*}{S2WAT~\cite{zhang2024s2wat}} 
                         & \multirow{1}{*}{Stylebank${^*}$~\cite{chen2017stylebank}} 
                        & \multirow{1}{*}{SaMST${^*}$ (Ours)} 
					\tabularnewline
					\hline 

					\multirow{1}{*}{ArtFID~\cite{wright2022artfid} $\downarrow$}
					 & 57.02 & 52.31 & 46.78
					& 58.10 & 53.97 & \underline{34.87} & 59.46 
					& 38.74 & 64.02 & \textbf{25.20}
					\tabularnewline

					\multirow{1}{*}{CF~\cite{wang2021evaluate} $\uparrow$}
					 & 0.320 & 0.363 & 0.475
					& 0.351 & 0.397 & \underline{0.515}  & 0.335 & 0.452 & 0.296 & \textbf{0.538}
					\tabularnewline

                     \multirow{1}{*}{GE +LP~\cite{wang2021evaluate} $\uparrow$}
					 & 1.125 & \underline{1.482} & 1.337
					& 1.295 & 1.184 & 1.265 & 1.312
					& 1.388 & 1.089 & \textbf{1.515}
					\tabularnewline

                        \hline

                        \multirow{1}{*}{FLOPs (G) $\downarrow$}
					 & 39.52 & 145.68 & 1283.45
					& 63.30 & 179.89 & 291.44 & \underline{11.06} 
					& 582.62 & 35.48 & \textbf{5.31}
					\tabularnewline

                        \multirow{1}{*}{Time (ms) $\downarrow$}
					 & 247.67 & 15.54 & 194.76
					& 14.92 & 33.36 & 24.01 & \underline{3.96}
					& 94.88 & 5.60 & \textbf{1.03}
					\tabularnewline

                    \multirow{1}{*}{Params (M) $\downarrow$}
					 & 2.42 &  62.83 &  35.39
					&  7.01 & 4.09 & 11.18 &  \textbf{0.47}
					& 64.96 & 590.79 & \underline{0.91}
					\tabularnewline

                    \multirow{1}{*}{OIP (M) $\downarrow$}
					 & 2.42 &  62.83 &  35.39
					&  7.01 & 4.09 & 11.18 &  \underline{0.47}
					& 64.96 & 1.97 & \textbf{0.11}
					\tabularnewline

                    \multirow{1}{*}{Style Capacity $\uparrow$}
					 & $\infty$ &  $\infty$ &  $\infty$
					&  $\infty$ & $\infty$ & $\infty$ &  $\infty$
					& $\infty$ & 500+ & 50k+
					\tabularnewline

					\hline
	
			\end{tabular}
   }
		\end{center}
  \label{comparisonwsotas}
  \vspace{-15pt}
	\end{table*}

\begin{figure}[t]
\centering
\begin{minipage}[t]{0.48\textwidth}
\centering
\includegraphics[width=5cm]{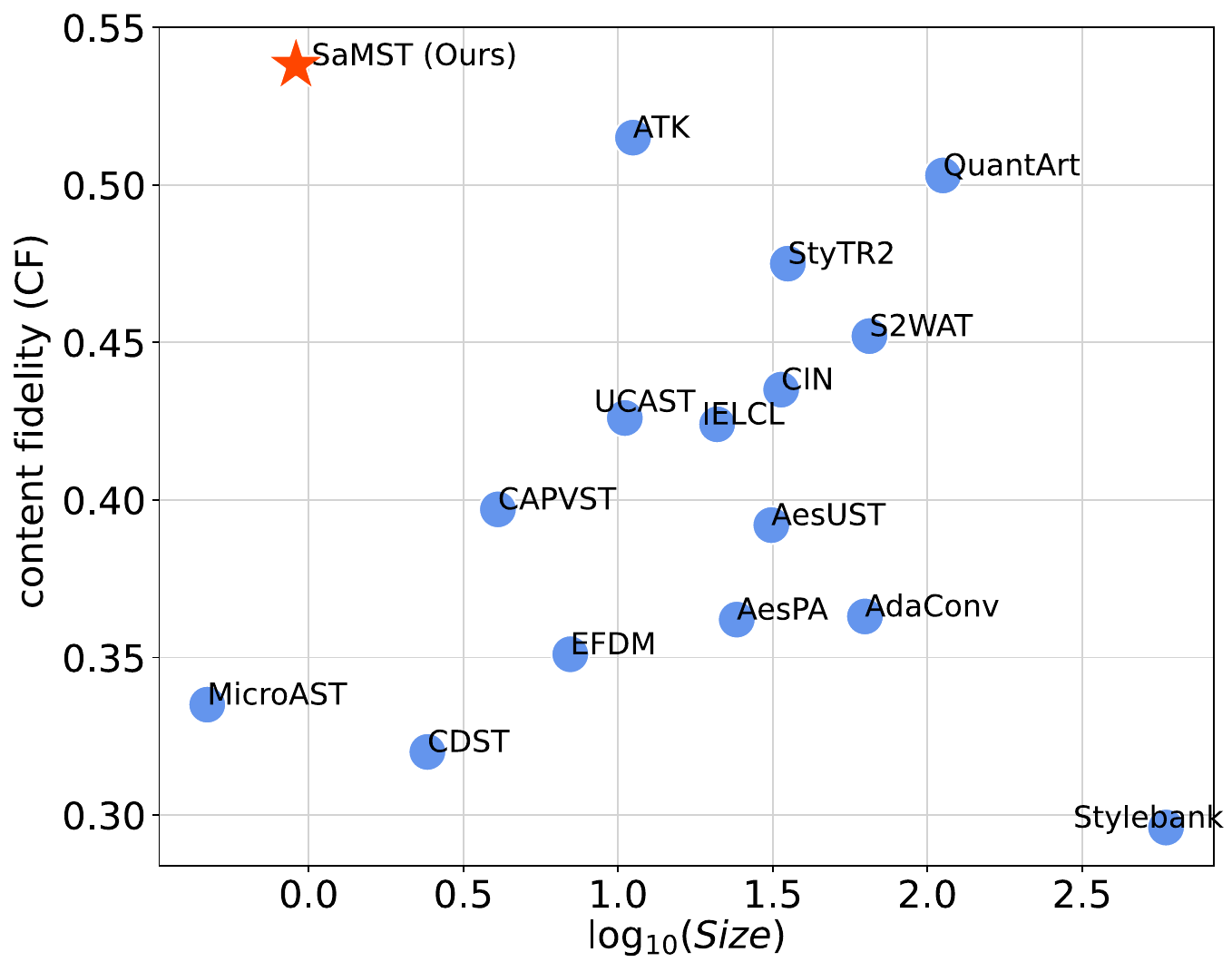}
\caption{Comparison of CF score \cite{wang2021evaluate} and model size (M).}
\label{scatter2}
\end{minipage}
\hspace{5pt}
\begin{minipage}[t]{0.48\textwidth}
\centering
\includegraphics[width=5cm]{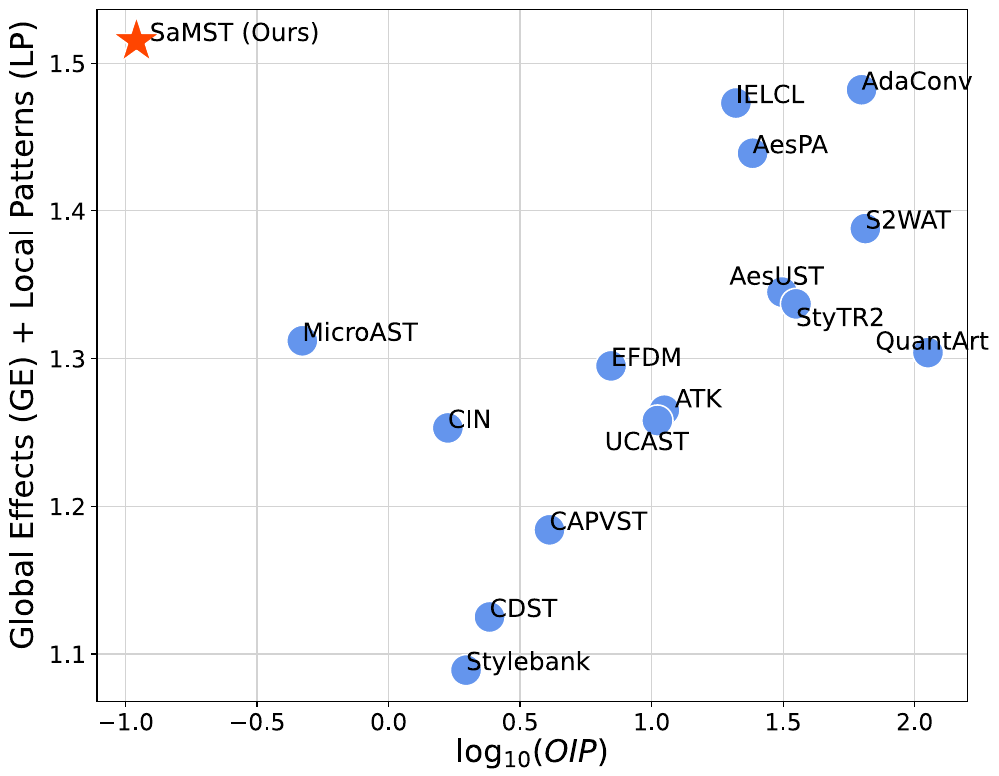}
\caption{Comparison of GE+LP score \cite{wang2021evaluate} and OIP (M).}
\label{scatter3}
\end{minipage}
\vspace{-20pt}
\end{figure}

\vspace{-15pt}
\subsubsection{Quantitative Comparison}
We resort to some quantitative metrics to better evaluate the proposed method. The results are shown in Table~\ref{comparisonwsotas}, Fig.~\ref{scatter1}, Fig.~\ref{scatter2} and Fig.~\ref{scatter3}. 

\noindent\textbf{(1) Stylization Quality:} Wright \emph{et al.}~\cite{wright2022artfid} propose ArtFID to measure stylized image quality from both style and content. Wang \emph{et al.}~\cite{wang2021evaluate} propose content fidelity (CF), global effects (GE) and local patterns (LP) to evaluate the quality of ST.  In detail, CF measures the faithfulness to content characteristics; GE assesses the stylization quality in terms of the global effects like global colors and holistic textures; LP assesses the stylization quality in terms of the similarity and diversity of the local style patterns. We collect 500 content images and 100 style images to synthesize $50k$ stylized images for each method and show their average metric scores in Table~\ref{comparisonwsotas}. Our SaMST achieves best results on the 3 metrics, indicating that it can transfer sufficient style patterns while better preserving the content details.

\noindent\textbf{(2) Efficiency and Application:} As shown in Table~\ref{comparisonwsotas} and Fig.~\ref{scatter1}, our SaMST achieves the best stylization results with the lowest computation quantity and inference time. This is because our style representation scheme contains enough style information compared with heavy VGG extractor. As for OIP (\emph{i.e.}, number of parameters for each style) in Table~\ref{comparisonwsotas}, our SaMST requires only $0.11M$ parameters while other methods suffer from parameter redundancy (\emph{e.g.}, $64.96M$ in S2WAT~\cite{zhang2024s2wat}). This indicates that our SaMST is easy to be deployed on edge devices or cloud servers, which has strong application value. Moreover, when adding new styles, size of Stylebank increases quickly ($1.18M$ per style), resulting in heavy storage burden. In contrast, size of SaMST increases slowly ($16$ per style). It is because that SCB in SaMST is capable of storing a large number of image styles in limited storage space. Then our SaMST can cover the need for various image styles in application scenarios.

\vspace{-10pt}
\subsection{Model Analysis}

\begin{table}[t]
		\caption{Ablation Study. Note that, we use CIN~\cite{dumoulin2016learned} with 3 resblocks as baseline for fair comparison (\emph{i.e.}, model 1). All model variants are trained on $50k$ style images. Runtime and FLOPs are evaluated on $512\times512$ images.}
 \vspace{-27pt}
		\label{tab0}
		\begin{center}
			\resizebox{1\hsize}{!}{
                    \scriptsize
				\begin{tabular}{|l|c|c|c|c|c|c|c|c|ccc|}
					\hline
					\multirow{2}{*}{Method} & \multirow{2}{*}{Param (M)} & \multirow{2}{*}{OIP (M)} & \multirow{2}{*}{Time (ms)}& \multirow{2}{*}{FLOPs (G)} & \multirow{2}{*}{Sconv} & \multirow{2}{*}{SRAdaIN} & \multirow{2}{*}{SCM} & \multirow{2}{*}{$\mathcal{L}_{geo}$} 
					& \multicolumn{3}{c|}{Metric}
					\tabularnewline
     
					&&&&&&&&& ArtFID$\downarrow$ & CF$\uparrow$ & GE+LP$\uparrow$
					\tabularnewline			
     
					\hline
					Model 1 & 54.58 & 0.18 & 4.08 & 7.16 & \faTimes &  \faTimes  & \faTimes 
                      & \faTimes  & 45.83 & 0.417 & 1.191
 					\tabularnewline
     
					Model 2  & 55.30 & 0.10 & 0.76 & 5.35 & \faCheck & \faTimes & \faTimes  
                      & \faTimes  & 38.62 & 0.382 & 1.306
					\tabularnewline
     
					Model 3 & 0.99 & 0.19 & 4.32 & 7.12 & \faTimes  & \faCheck & \faTimes  
                       & \faTimes  & 47.89 & \textbf{0.594} & 1.214
					\tabularnewline

                        Model 4 & 0.91 & 0.11 & 0.98 & 5.31 & \faCheck  & \faCheck & \faTimes  
                       & \faTimes  & 33.24 & 0.467 & 1.408
					\tabularnewline

                        Model 5 & 0.91 & 0.11 & 1.03 & 5.31 & \faCheck & \faCheck & \faCheck  
                      &\faTimes  & 28.36 & 0.486 & 1.464
                      \tabularnewline
     
					Model 6 (Ours) & 0.91 & 0.11 & 1.03 & 5.31 & \faCheck & \faCheck & \faCheck  
                      & \faCheck  & \textbf{25.20} & 0.528 & \textbf{1.515}
					\tabularnewline

					\hline	
			\end{tabular}}
		\end{center}
  \label{ablationall}
  \vspace{-10pt}
	\end{table}

\begin{figure}[t]
		\centering
    \includegraphics[width=1\linewidth]{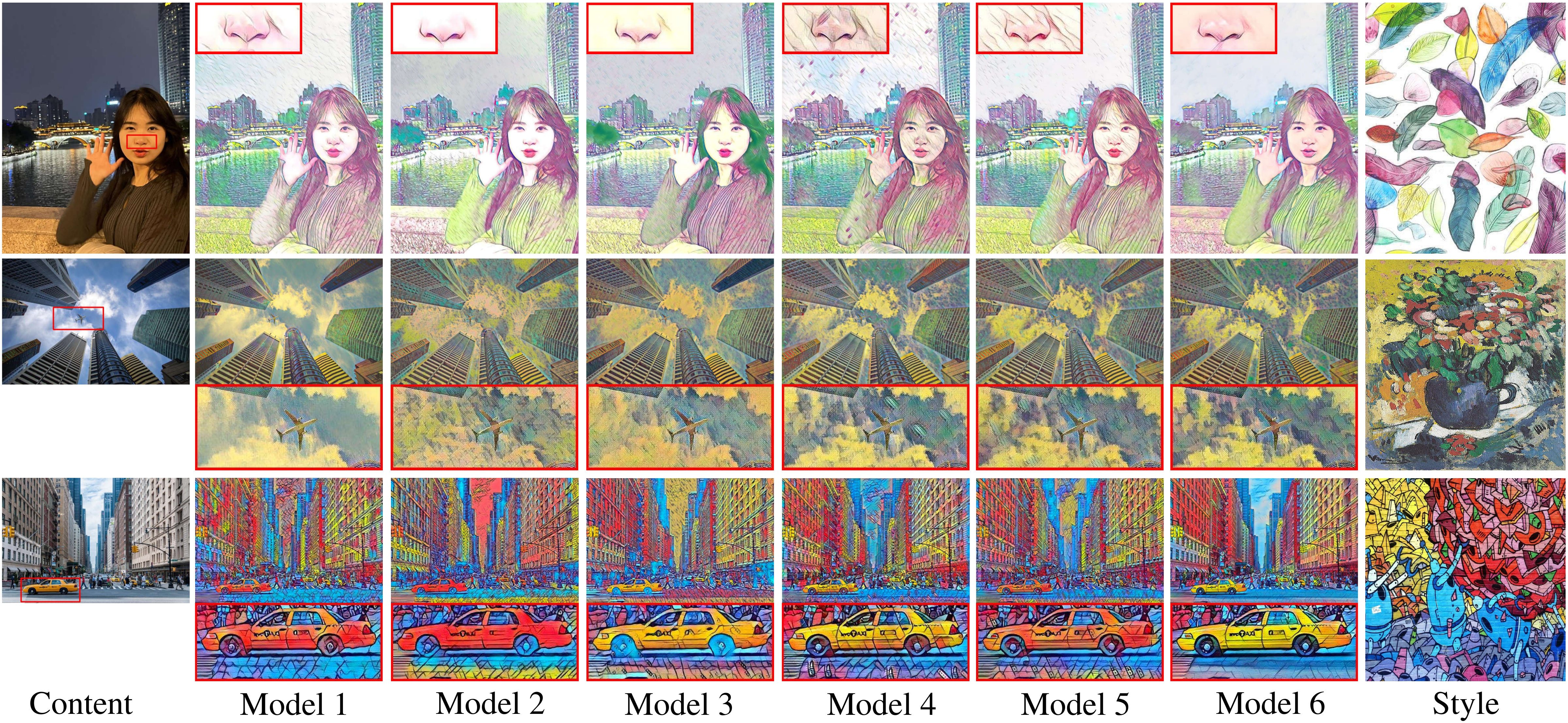}
  \vspace{-15pt}
		\caption{Ablation study of SaMST. We also display the zoomed patch in the red box for better evaluation. The settings of the model variants are shown in Table~\ref{ablationall}.}
    \vspace{-18pt}
 \label{visual_ablation}
\end{figure}

We demonstrate the effectiveness of our proposed components in this section. And more model analysis can be found in supplemental material.

\noindent\textbf{(1) SConv:} We propose SConv to reproduce local geometric structures from style to content. To validate its effectiveness, we introduce a model variant (model 1 in Table~\ref{ablationall}) by removing all designs. In generator, we replace the common convolution layer by our SConv layer to obtain model 2. Although model 2 achieves better inference time, model size increases more quickly. This is because both style representations in SCB and learnable variables in conditional IN layer grow with the number of styles. Moreover, SConv helps achieve significantly higher ArtFID and GE+LP score than model 1. From the second scene in Fig.~\ref{visual_ablation}, model 2 produces image textures that closed to the style image.

\noindent\textbf{(2) SRAdaIN:} To make stylized images capture global properties from style images, SRAdaIN is employed in our method. To demonstrate its effectiveness, we add SRAdaIN to obtain model 1 to obtain model 3 for comparison. It can be observed from Table~\ref{ablationall} that the SRAdaIN improves CF and GE+LP score, which indicates that style representations contains more accurate global style properties. Moreover, the model size is much smaller than model 1. In addition, we further develop model 4 by adding SConv to model 3. With both SConv and SRAdaIN, model 4 produces significant higher metrics as compared to model 1. From the first scene in Fig.~\ref{visual_ablation}, model 4 produces soft tones similar with the style image. Moreover, from the third scene in Fig.~\ref{visual_ablation}, model 4 produces sharper edges of higher perceptual quality.

\begin{figure}[t]
		\centering
    \includegraphics[width=0.75\linewidth]{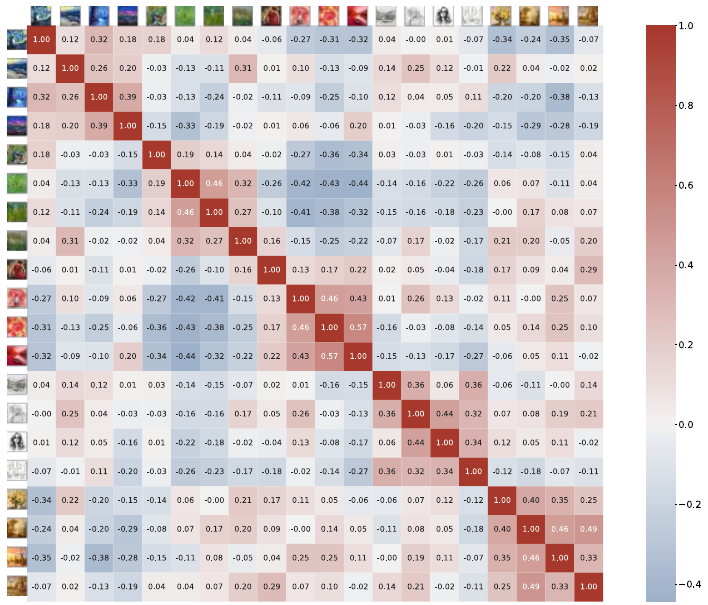}
    \vspace{-7pt}
    \caption{Style correlation matrix.}
    \vspace{-25pt}
 \label{correlation_matrix}
\end{figure}

\noindent\textbf{(3) SCM:} Although model 4 produces soft tones, it also produces artifacts on stylized images (\emph{e.g.}, the second scene in Fig.~\ref{visual_ablation}). As we introduce SCM to model 4, we obtain model 5 to achieve better visual quality. And model 5 achieves better CF score, which indicates that stylized images suffer from less image distortion.

\noindent\textbf{(4) Geometric Consistency Loss:} The geometric consistency loss is introduced to preserve scene geometry. The loss $\mathcal{L}_{geo}$ is employed in our method. We add $\mathcal{L}_{geo}$ to model 5 to obtain model 6 for comparison. Model 6 achieves best ArtFID and GE+LP score and second best CF score. It demonstrates that model 6 achieves the better balance between style transformation and content preservation. As shown in the first and third scene of Fig.~\ref{visual_ablation}, model 6 produces clearer stylized results and image details. In contrast, models trained without $\mathcal{L}_{geo}$ produce messy stylized image details.

\noindent\textbf{(5) Style Correlation:} Our method contains style information in compact representations. The representations are correlated with the visualization of style images. To demonstrate this, we randomly select 20 styles and corresponding style representations. We calculate correlation matrix of the selected styles. Then we visualize the result in Fig.~\ref{correlation_matrix}. The correlation matrix shows that the visually similar styles have higher correlation coefficient (\emph{e.g.}, high correlation coefficient between the sketch-wise styles). The correlation matrix provides strong evidence that the style representations distinguish styles in semantic and visual similarity.

\begin{figure}[t]
		\centering
    \includegraphics[width=0.9\linewidth]{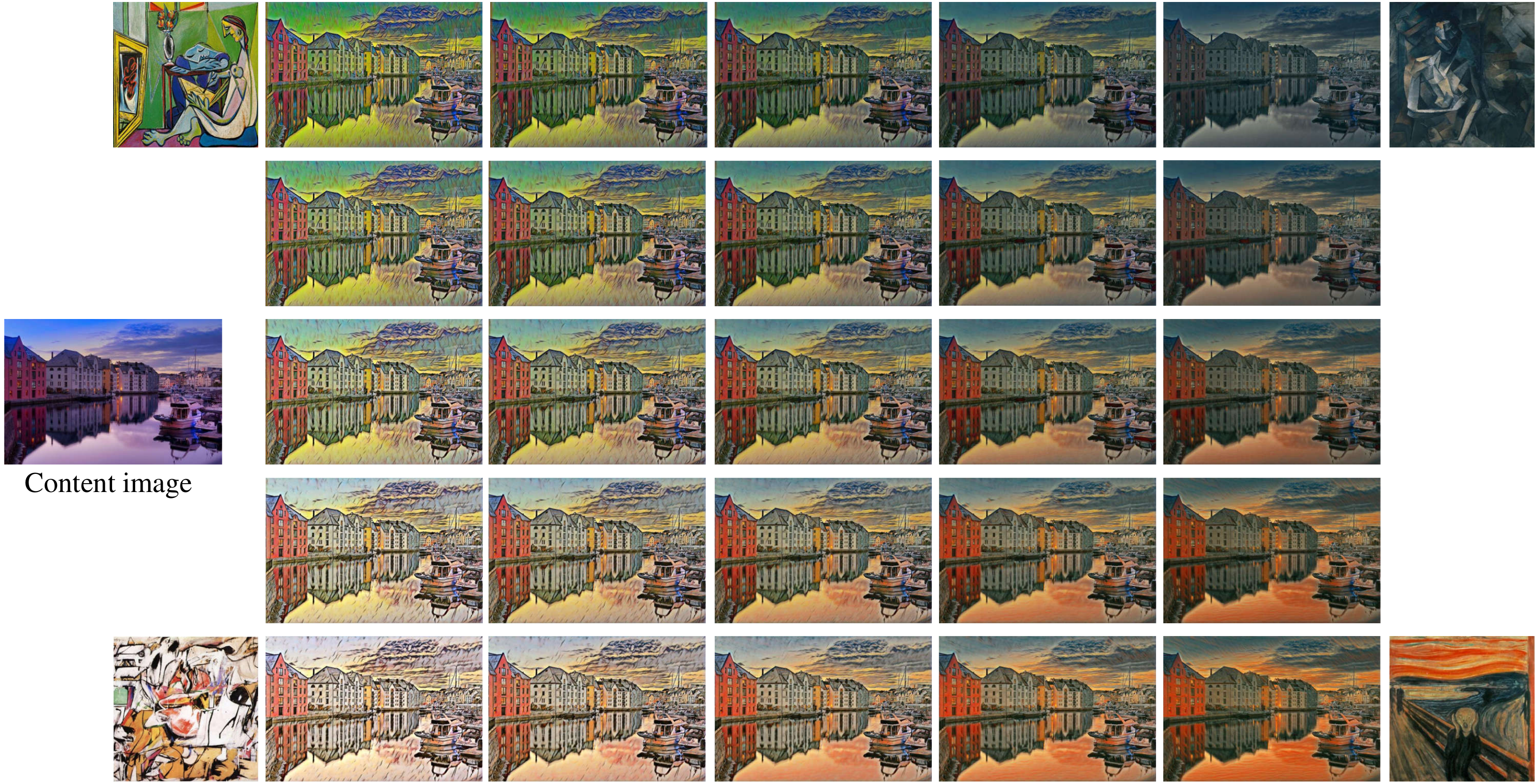}
    \vspace{-10pt}
    \caption{Visualization results by style interpolation.}
 \label{style_fusion1}
\end{figure}

\begin{figure}[t]
		\centering
    \includegraphics[width=1\linewidth]{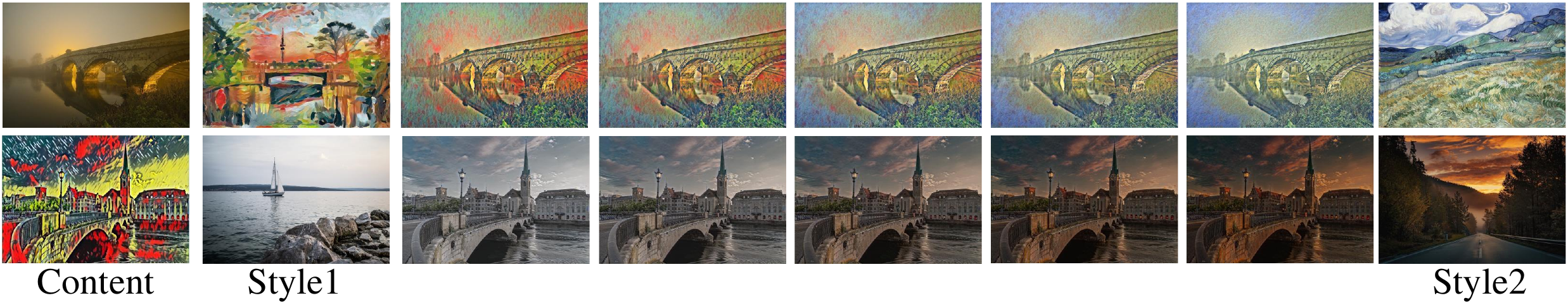}
    \vspace{-15pt}
    \caption{Visualization results by style interpolation. We also show the conversion between photorealistic images and artistic images.}
    \vspace{-15pt}
 \label{style_fusion2}
\end{figure}

\vspace{-10pt}
\subsection{User Interaction}

\noindent\textbf{(1) Style Interpolation:} To blend a set of $K$ styles $s_1,s_2,...,s_K$, we interpolate between style representations $f_1,f_2,...,f_K$ to obtain overall style representation $f_{mixed}=\sum_{i=1}^K{{w_i}{f_i}}$ such that $\sum_{i=1}^K{w_i}=1$. Then the stylized image is generated:
\begin{align}
I_{mixed}={\rm SaMST}(c,f_{mixed}).
\end{align}
The results are shown in Fig.~\ref{style_fusion1} and Fig.~\ref{style_fusion2}. Figure~\ref{style_fusion1} illustrates the results interpolated with the different weight levels. The style pattern gradually shift to other style patterns by changing the interpolation weights. As shown in Fig.~\ref{style_fusion2}, our SaMST can converts between photorealistic images and artistic images (\emph{e.g.}, photorealistic content to artistic styles in $1^{st}$ row and artistic content to photorealistic styles in $2^{nd}$ row).

\begin{figure}[t]
		\centering
    \includegraphics[width=1\linewidth]{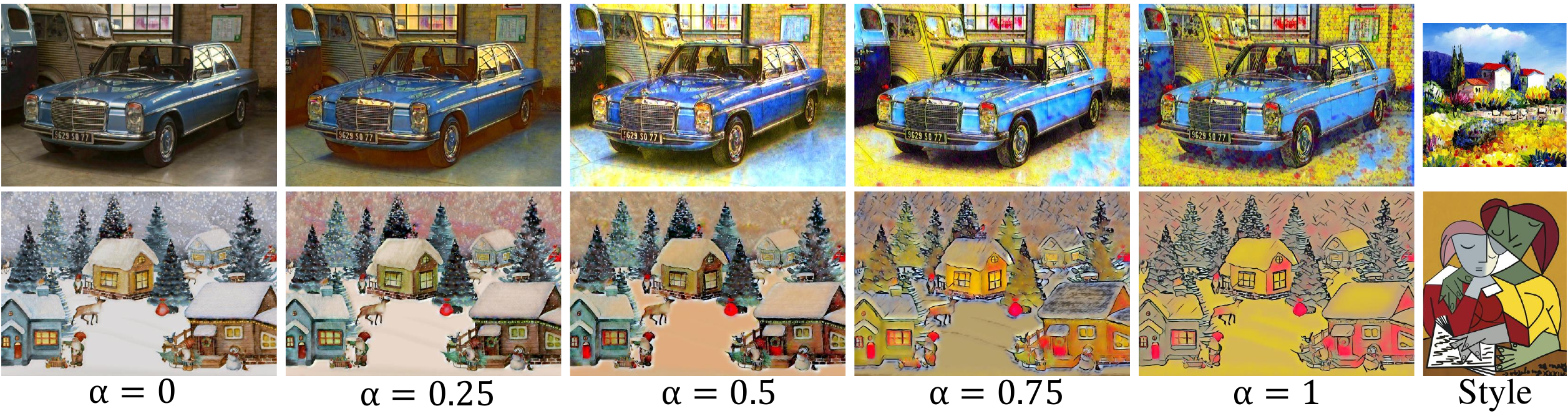}
    \vspace{-18pt}
    \caption{Content-style trade-off. We can control the balance between content and style by changing the factor $\alpha$ in Eq.~\ref{cs_tradeoff}.}
    \vspace{-9pt}
 \label{tradeoff}
\end{figure}

The degree of style transfer can be controlled during training by adjusting the style weight $\lambda_s$ in Eq.~\ref{loss_fun}. Our SaMST allows an alternative to implement content-style tradeoff:
\begin{align}
\label{cs_tradeoff}
I_{\alpha{i}}={\rm SaMST}(c,(1-\alpha)f_0 + \alpha{f_i}),
\end{align}
where $\alpha$ is the degree factor to control the stylized images (as shown in Fig.~\ref{tradeoff}).

\begin{figure}[t]
		\centering
    \includegraphics[width=0.8\linewidth]{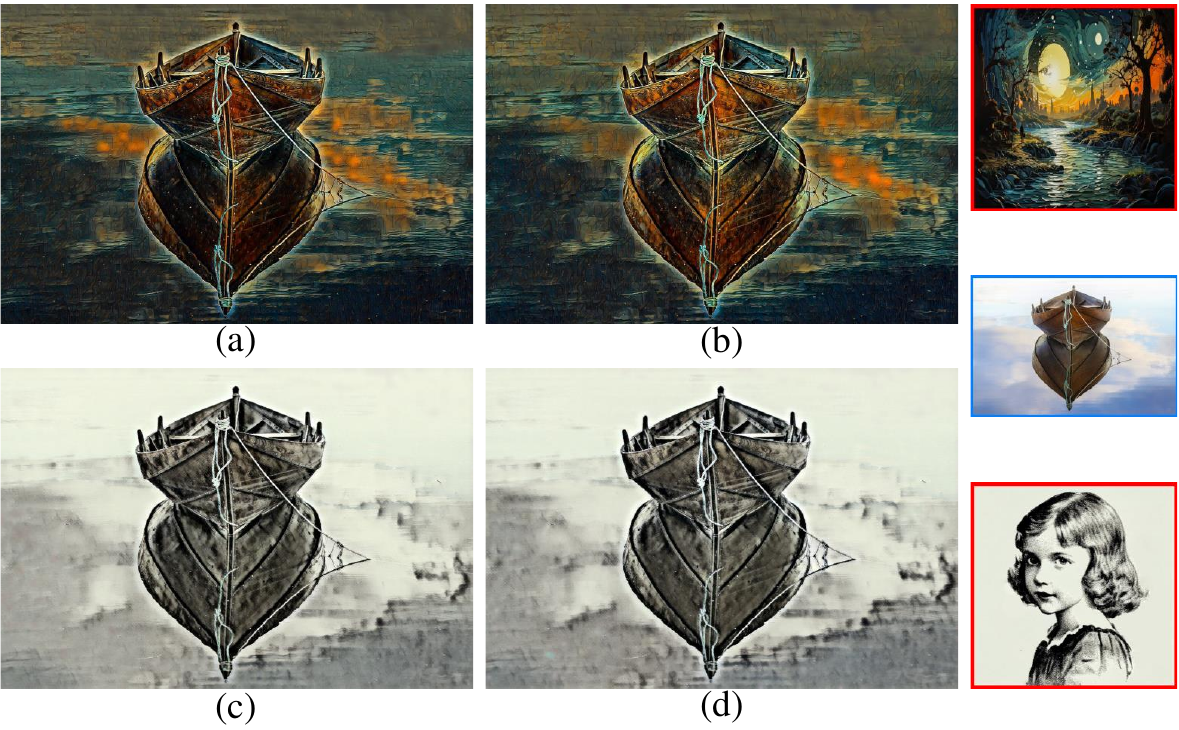}
    \vspace{-10pt}
    \caption{Comparison between incremental training (a)(c) and fresh training (b)(d). The content and styles are shown on the right.}
    \vspace{-18pt}
 \label{style_incremental}
\end{figure}

\noindent\textbf{(2) Incremental Training:} We enable an incremental training for new styles, which has comparable learning time to the online-learning method~\cite{gatys2015neural}, while model size grows slower than previous method~\cite{chen2017stylebank}. Specifically, we fix model 6 in Table~\ref{ablationall}, then train the style representations for new styles (as shown in Algorithm~\ref{algorithm_incre_training}). The style converges very fast since only the 16-dimension style representation would be updated in iterations instead of the whole model. In our experiments, when training with NVIDIA RTX 3090 GPU and given training image size of 256, it only takes around 1 minute with about $3k$ iterations to train a new style.  Figure~\ref{style_incremental} shows stylized results of new styles by incremental training. Compared to fresh training (\emph{i.e.}, retraining the whole network with the new styles), our learning scheme obtain very similar stylized results. More incremental training results can be found in supplemental material.

\vspace{-10pt}
\section{Conclusion}
\vspace{-10pt}
In this paper, we propose a style representation learning scheme to store accurate style information. Moreover, we introduce a lightweight and practical style-aware multi-style transfer (SaMST) network to achieves efficient ST. In addition, we propose a incremental training scheme to expand new styles without forgetting. It is demonstrated that our style representation learning scheme can extract accurate and robust style information. Experimental results show that our network achieves state-of-the-art performance for ST task.

\textbf{Acknowledgement.} This work was partially supported by the Guangdong Basic and Applied Basic Research Foundation (2022B1515020103), and the Shenzhen Science and Technology Program (No. RCYX20200714114641140).

%
%
\bibliographystyle{splncs04}
\bibliography{main}
\end{document}